\documentclass[conference]{IEEEtran}
\IEEEoverridecommandlockouts

\usepackage{algorithm}
\usepackage{algorithmic}
\usepackage{amsmath}
\usepackage{booktabs}
\usepackage{tabularx}
\usepackage{amssymb}
\usepackage[numbers,sort&compress]{natbib}
\usepackage{url}
\usepackage{graphicx}
\usepackage{svg}
\usepackage{multirow}
\usepackage[table]{xcolor}
\usepackage{tikz}
\usepackage{soul}
\usepackage{enumitem}
\usepackage{pgfplots}
\pgfplotsset{compat=1.18}
\usepackage{subcaption}
\usepackage{adjustbox}
\usepackage{caption}
\usepackage{lipsum}
\usepgfplotslibrary{groupplots}  
\usepackage{tcolorbox}

\usetikzlibrary{external}
\usepgfplotslibrary{statistics}
\def\BibTeX{{\rm B\kern-.05em{\sc i\kern-.025em b}\kern-.08em
    T\kern-.1667em\lower.7ex\hbox{E}\kern-.125emX}}
\begin{document}

\title{Abstention vs. Hallucination: Benchmarking LLM Source Attribution for Scientific Citations}

\author{
\IEEEauthorblockN{
Deepa Tilwani\IEEEauthorrefmark{1},
Yash Saxena\IEEEauthorrefmark{2},
Seyedali Mohammadi\IEEEauthorrefmark{2},
Ankur Padia\IEEEauthorrefmark{2},
Edward Raff\IEEEauthorrefmark{2},
Amit Sheth\IEEEauthorrefmark{1}\IEEEauthorrefmark{3},\\
Srinivasan Parthasarathy\IEEEauthorrefmark{4},
Manas Gaur\IEEEauthorrefmark{2}
}

\IEEEauthorblockA{
\IEEEauthorrefmark{1}
Artificial Intelligence Institute, University of South Carolina,
Columbia, South Carolina, USA
}

\IEEEauthorblockA{
\IEEEauthorrefmark{2}
University of Maryland, Baltimore County,
Baltimore, Maryland, USA
}

\IEEEauthorblockA{
\IEEEauthorrefmark{3}
IAIRO, India, 
\IEEEauthorrefmark{4}
The Ohio State University,
Columbus, Ohio, USA
}

\IEEEauthorblockA{
Emails:
dtilwani@mailbox.sc.edu,
ysaxena1@umbc.edu,
m294@umbc.edu,
pankur1@umbc.edu,\\
edraff1@umbc.edu,
amit@sc.edu,
srini@cse.ohio-state.edu,
manas@umbc.edu
}
}

\maketitle

\begin{abstract}
Large language models (LLMs) increasingly generate citation-backed responses, yet citation hallucination remains a major challenge for trustworthy scientific information access. We introduce \textsc{REASONS}, a benchmark of 12,723 sentence-level citation instances spanning 12 arXiv subject categories, designed to evaluate scientific citation attribution under varying evidence conditions. We propose a dual-metric framework consisting of Abstention Rate (AR) and Hallucination Rate (HR) to characterize the trade-off between reliability and responsiveness. Using author-attribution and title-attribution tasks, we evaluate proprietary and open-source LLMs under zero-context, metadata-augmented, cascaded metadata-augmented prompting (CMP), retrieval-augmented, and adversarial settings. Advanced RAG lowers HR relative to Na\"ive RAG
(65.4\% vs. 87.6\%) but reduces AR from 5.0\% to 0\%. Under adversarial metadata, several systems exceed 85\% HR, while
retrieval-augmented variants frequently maintain near-zero abstention. Human evaluation of 1,000 outputs ($\kappa=0.78$) finds a
12.7:1 ratio of factual hallucinations to acceptable paraphrases. Our findings demonstrate that citation attribution systems should be evaluated not only for correctness but also for their ability to abstain appropriately under uncertainty. REASONS provides a benchmark and evaluation framework for studying attribution reliability in citation generation.

\end{abstract}


\section{Introduction}
\begin{figure*}[t]
\centering
\includegraphics[width=\textwidth, trim=0.2cm 0.2cm 0.2cm 0.5cm,
    clip]{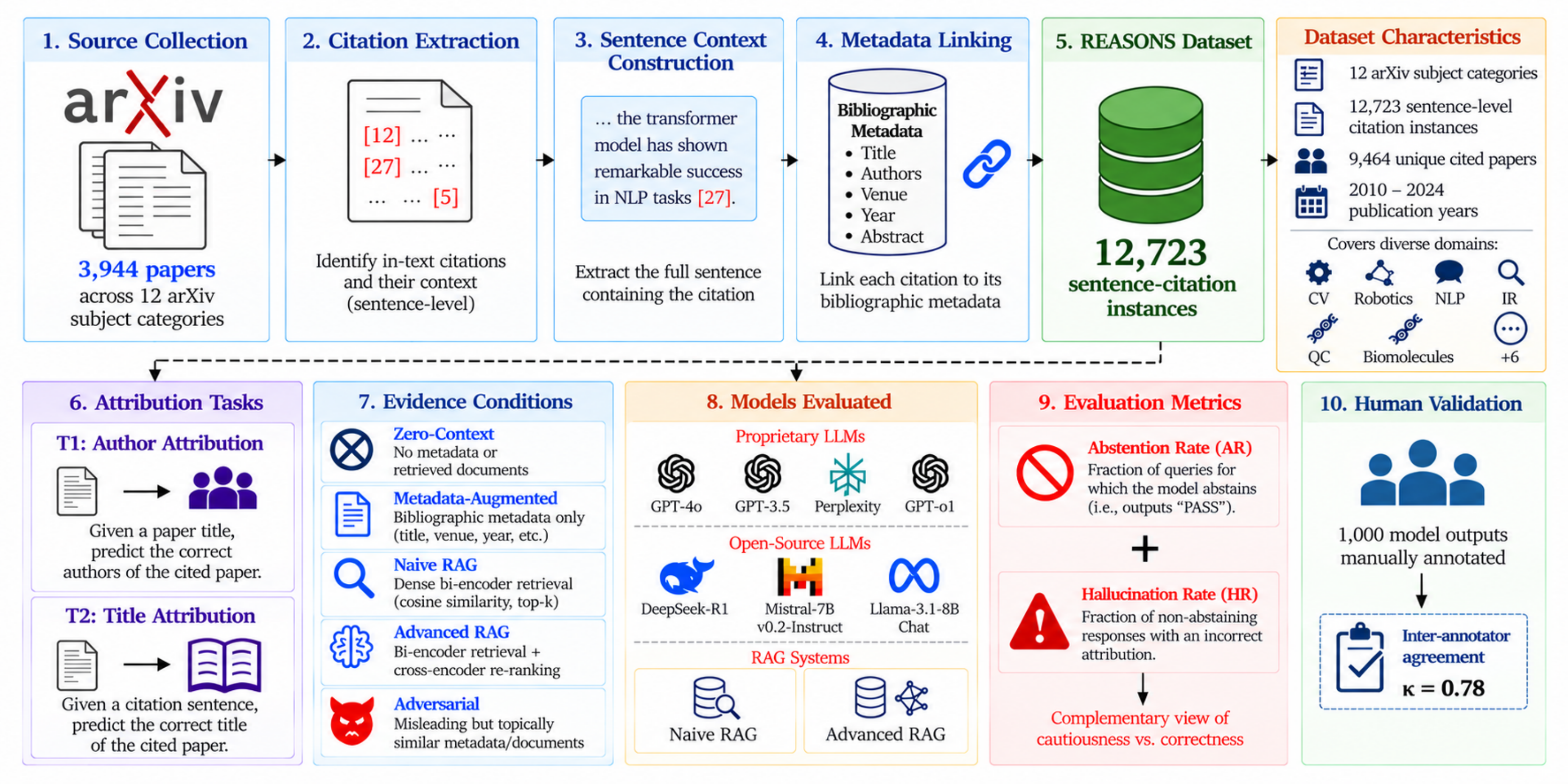}
\caption{
\footnotesize Overview of the REASONS benchmark construction and evaluation framework.}
\label{fig:reasons_framework}
\end{figure*}

Large language models (LLMs) increasingly generate citations to support their claims, yet citation hallucination poses a growing challenge to trustworthy scientific information access. Prior studies have documented fabricated and erroneous bibliographic citations in ChatGPT-generated responses~\cite{walters2023fabrication}, while more than 100 suspected citation hallucinations have recently been reported in NeurIPS 2025 accepted papers~\cite{ShmatkoAdamEsau2026NeurIPS}. Recent theoretical work further argues that hallucination cannot be completely eliminated under general learning-theoretic conditions~\cite{xu2024hallucination}. Existing work has investigated mechanisms and evaluation strategies for LLM source attribution and citation generation~\cite{tilwani2024neurosymbolic,saxena2025generation}. However, a critical behavioral question remains underexplored: \textit{when generating citations through parametric knowledge or retrieval, do systems abstain when evidence is insufficient, or do they confidently produce incorrect attributions?}
As citation-aware systems like Perplexity~\cite{roose2024can} and Bing~\cite{mehdi2023reinventing} increasingly mediate scientific discovery, understanding attribution behavior has become essential. Figure~\ref{fig:reasons_framework} previews the overview of the REASONS benchmark construction and evaluation framework.

Prior work optimizes citation accuracy via better retrieval~\cite{han2025retrieval,lewis2020retrieval} or training~\cite{chuang2025selfcite, khalifasource}, but ignores \textit{when and how often systems should abstain rather than hallucinate}. Current benchmarks fail in three ways: paragraph-level evaluation masks claim-level errors~\cite{bohnet2022attributed,gao-etal-2023-enabling}, Wikipedia datasets suffer LLM contamination~\cite{petroni2020kilt,wagner2025}, and metrics ignore epistemic safety~\cite{lin2022truthfulqa}. Critically, existing work cannot answer: \textit{how frequently do systems abstain vs. hallucinate under varying evidence conditions}, and \textit{does adding context shift behavior from cautious refusal to confident error?} Scientific citations offer sentence-level attributions with verifiable ground truth, minimal contamination, and domain variation, enabling controlled measurement of abstention-hallucination tradeoffs as evidence availability changes. We make the following contributions: 

\begin{enumerate}[itemsep=0pt, leftmargin=*]

\item We introduce REASONS, the first sentence-level citation benchmark designed for behavioral evaluation, comprising 12,723 citations across 12 arXiv subject categories that enable systematic measurement of abstention and hallucination rates—metrics absent from existing attribution benchmarks.

\item We propose a dual-metric evaluation framework that jointly measures \textit{abstention rate} (AR) and \textit{hallucination rate} (HR) to characterize epistemic safety in citation generation. We demonstrate that standard attribution improvements create an abstention-hallucination tradeoff where increased responsiveness eliminates epistemic caution.

\item We validate our automated metrics through stratified human evaluation of 1,000 model outputs across five systems and multiple domains, demonstrating strong correlation ($r=0.89$, $p<0.001$) between automated HR and human-judged attribution errors. Our analysis confirms that hallucination is predominantly a factual error problem (12.7:1 ratio of factual errors to acceptable paraphrases) rather than a paraphrase detection issue, and reveals that systems struggle to calibrate abstention decisions appropriately.

\end{enumerate}


\section{Background and Related Work}

\noindent Attribution optimization in LLMs has focused primarily on improving citation correctness through better retrieval or training. RETRO~\citep{borgeaud2022improving} retrieves text chunks during pretraining to ground generation in external knowledge, while retrieval-augmented generation (RAG)~\citep{lewis2020retrieval} combines dense retrieval with sequence-to-sequence models for knowledge-intensive tasks. Source-aware training methods~\citep{khalifasource, huang-etal-2024-advancing} teach models to associate generated content with source identifiers, and self-supervised alignment techniques~\citep{chuang2025selfcite} refine citation placement through contrastive learning. Attributed QA~\citep{bohnet2022attributed} evaluates whether model outputs are supported by retrieved passages. While these approaches demonstrate that retrieval and training can improve citation accuracy, they evaluate whether systems \textit{can} cite correctly when evidence exists, not whether systems \textit{should} abstain when evidence is insufficient or misleading. This leaves unanswered the behavioral question raised in our introduction: when generating citations, do LLMs know when they don't know, or do they confidently fabricate sources despite uncertainty? 

Evaluation frameworks for attribution have largely focused on verifying whether generated content is supported by retrieved evidence. Benchmarks such as Attributed Question Answering (AQA) \citep{bohnet2022attributed},
ALCE \citep{gao-etal-2023-enabling}, and related knowledge-grounding and
fact-verification benchmarks \citep{petroni2020kilt, thorne-etal-2018-fever}
assess whether generated responses or claims are supported by relevant evidence. Similarly, citation-intent datasets and large scholarly corpora
\citep{cohan2019structural, lo2020s2orc} support citation analysis,
scholarly document understanding, and modeling of citation relationships. Although these benchmarks provide important measures of citation accuracy and retrieval effectiveness, they primarily evaluate whether systems \emph{can} cite correctly when evidence is present. Existing evaluation metrics are also based on precision, recall, or support-based scoring, which cannot distinguish systems that achieve high precision by abstaining from uncertain cases from those that respond to all queries while producing confident attribution errors. Consequently, these frameworks do not capture the fundamental behavioral tradeoff between abstention and hallucination, nor do they examine how models behave when evidence is incomplete, ambiguous, or misleading—conditions that frequently arise in real-world knowledge-seeking settings. Table~\ref{tab:dataset_comparison} situates \textsc{REASONS} relative to existing datasets and shows that it is the first benchmark designed to systematically measure abstention–hallucination tradeoffs in the generation of citations. 

Recent work on adversarial retrieval~\citep{abolghasemi2024} reveals that LLMs exhibit attribution bias when retrieved passages are topically relevant but do not support specific claims, exposing a critical gap between retrieval quality (finding documents with high semantic similarity) and attribution safety (knowing when retrieved evidence is insufficient for the specific claim being made). Unlike search engines that present ranked results for user inspection, generative RAG pipelines convert retrieved evidence directly into assertions, raising the possibility that surface similarity might trigger confident citation regardless of claim-level support. This suggests that retrieval might make systems more confident without making them more correct, directly relevant to our question of whether adding evidence causes systems to shift from saying ``\textit{I don't know}" to asserting wrong answers with high confidence.

\noindent Domain adaptation studies~\citep{gururangan2020don} show that task performance varies across domains, with general-purpose LLMs exhibiting stronger capabilities in well-represented areas such as Computer Vision than in specialized fields such as Quantum Computing, due to differences in pretraining corpus composition. However, whether attribution safety, the joint behavior of abstention and hallucination, degrades systematically with domain complexity or varies stochastically with model capacity remains unknown. This gap is critical for deployment: systems may perform well on aggregate benchmarks dominated by popular domains, yet exhibit unsafe behavior on specialized queries where epistemic caution is most critical. This concern is particularly relevant in high-stakes domains, where
reliability and safety are central requirements for trustworthy AI
deployment~\citep{tilwani2026neurosymbolic}. Existing work leaves our behavioral questions unanswered. 
Sentence-level granularity is not perfect: some claims require multi-sentence discourse, tables/figures, or broader context to verify. Our evaluation settings explicitly vary the amount of context available (e.g., minimal vs. metadata/retrieval conditions), allowing us to study how systems trade off abstention and hallucination as evidence becomes more complete. Sentence-level evaluation provides a controlled lens for observing attribution behavior under systematically manipulated evidence.

\begin{table}[t]
\centering
\small
\resizebox{\columnwidth}{!}{
\begin{tabular}{@{}lcccc@{}}
\toprule
\textbf{Dataset} & \textbf{Granularity} & \textbf{Domains} & \textbf{Abstention} & \textbf{Task} \\
\midrule
S2ORC~\citep{lo2020s2orc} & Document & General & No &  Scholarly Corpus \\
UnarXive~\citep{saier2020unarxive} & Section & CS & No & Citation Analysis \\
RefSum~\citep{liu-etal-2021-refsum} & Paragraph & CS & No & Summarization \\
SciCite ~\citep{cohan2019structural} & Sentence & Comp. Ling. & No & Citation Intent Classification \\
\midrule
\textsc{REASONS} (ours) & Sentence & 12 research areas & \textbf{Yes} & \textbf{Attribution} \\
\bottomrule
\end{tabular}}
\caption{\footnotesize Comparison of \textsc{REASONS} with existing citation datasets.}
\label{tab:dataset_comparison}
\end{table}

\section{Dataset Construction and Benchmark}

\begin{table}[t!]
\footnotesize
\centering
\begin{tabular}{p{3.5cm}p{1cm}p{1.5cm}p{1cm}}
\toprule
\textbf{Domain} & \textbf{Paper Count} & \textbf{Papers in IEEE format} & \textbf{Citation Count} \\
\midrule
Computer Vision (CV) & 5488 & 1028 & 3437 \\ 
Robotics & 3656 & 292 & 776 \\
Graphics & 1796 & 384 & 1417 \\
Information Retrieval (IR) & 1741 & 564 & 1654 \\
Artificial Intelligence (AI) & 1697 & 531 & 2021 \\
Natural Language Processing (NLP) & 1526 & 293 & 1092 \\
Cryptography & 1084 & 371 & 1106 \\
Neurons and Cognition (NNC) & 892 & 111 & 326 \\
Human Computer Interaction (HCI) & 761 & 112 & 229 \\
Databases & 723 & 115 & 182 \\
Quantum Computing (QC) & 421 & 126 & 456 \\ 
Biomolecules & 119 & 17 & 27 \\
\midrule[1pt]
\textbf{Total} & \textbf{19904} & \textbf{3944} & \textbf{12723} \\ \bottomrule
\end{tabular}
\caption{\footnotesize Our benchmark dataset, \textsc{REASONS}, includes papers and sentences from 12 arXiv subject categories }
\label{table:domain-summary}
\end{table}
\begin{table}[t]
\centering
\footnotesize
\setlength{\tabcolsep}{4pt}
\begin{tabularx}{\columnwidth}{p{2.2cm} X}
\toprule
\multicolumn{2}{c}{\textbf{\textsc{REASONS} dataset example (Computer Vision)}} \\
\midrule
Link & \url{http://arxiv.org/abs/2012.05435v2} \\
Paper title & \textit{Optimization-Inspired Learning with Architecture Augmentations and Control Mechanisms for Low-Level Vision} \\
Sentence ID & 32 \\
Sentence & We adopt the $\ell_1$-norm. For GM, we establish a residual network with seven convolutional layers and six ReLU blocks, each placed after a convolutional layer. The DM is constructed as a standard CNN-based classifier. \\
Citation text & C. Ledig et al., ``Photo-Realistic Single Image Super-Resolution using a Generative Adversarial Network,'' CVPR 2017. \\
Cited paper id & \texttt{arXiv:1609.04802} \\
Cited paper title & \textit{Photo-Realistic Single Image Super-Resolution Using a Generative Adversarial Network} \\
Abstract (excerpt) & Despite breakthroughs in accuracy and speed for single-image super-resolution using deep convolutional neural networks, ... \\
Authors & Christian Ledig, Lucas Theis, Ferenc Huszár, Jose Caballero, Andrew Cunningham, Alejandro Acosta, Andrew Aitken, Alykhan Tejani, Johannes Totz, Zehan Wang, Wenzhe Shi \\
\bottomrule
\end{tabularx}
\caption{\footnotesize Example instance from \textsc{REASONS} showing citation context and linked metadata.}
\label{tab:datapoint}
\end{table}

As a central contribution, REASONS (REtrieval and Automated
citationS Of scieNtific Sentences) is a sentence-level citation
attribution benchmark comprising 12,723 citation instances extracted
from 3,944 arXiv papers spanning 12 subject categories. Rather than artificially balancing domain representation, the dataset preserves the natural long-tail distribution (Table~\ref{table:domain-summary}), enabling systematic analysis of how attribution reliability varies between well-represented and highly specialized fields. This design allows us to disentangle whether attribution failures arise from domain specialization or simply reflect differences in data availability. Each instance contains a structured attribution datapoint linking the citing sentence to verified citation metadata, including the cited paper’s title, authors, abstract, and identifier (Table~\ref{tab:datapoint}). In addition to domain-level diversity, \textsc{REASONS} includes cross-domain citation markers that capture interdisciplinary references, providing controlled test cases in which topical similarity between citing and cited papers is reduced. These features enable fine-grained evaluation of attribution behavior under varying levels of semantic overlap and evidence availability. Link for dataset \url{https://zenodo.org/records/18264033}.


\noindent \textit{Metadata enrichment and structure:} Each instance pairs a citing sentence with verified metadata of the referenced paper, including title, abstract, author list, venue, and publication year. This metadata is verified by exact matching of in-text citation strings
to retrieved records. The source explicitly cited by the original
authors is treated as ground truth; thus, REASONS measures fidelity
to the observed citation rather than whether alternative sources could
also support the claim. The dataset also records the source paper
identifier and publication date, enabling analyses of temporal factors
such as model knowledge cutoffs.



\paragraph{\textbf{Tasks}}
We evaluate attribution behavior through two complementary tasks that probe different aspects of citation knowledge. \textit{Task T1 (Direct Attribution)} provides the model with a paper title and asks it to return the author list or \texttt{Abstain} if uncertain. This tests whether models can ground metadata when the source identity is explicitly given, isolating their ability to recall or verify authorship information. \textit{Task T2 (Indirect Attribution)} provides a citing sentence with an inline citation placeholder and asks the model to identify the cited paper's title or abstain. This simulates deployed settings in which models must infer which source supports a claim without author-provided citations, thereby testing attribution behavior when source identity must be determined. These tasks separate memorization of paper-author mappings (T1) from claim-to-source attribution reasoning (T2) (prompts in the supplementary).

\begin{figure}[t]
  \centering
  \includegraphics[width=0.5\textwidth, trim = 1cm 1cm 4cm .5cm, clip]{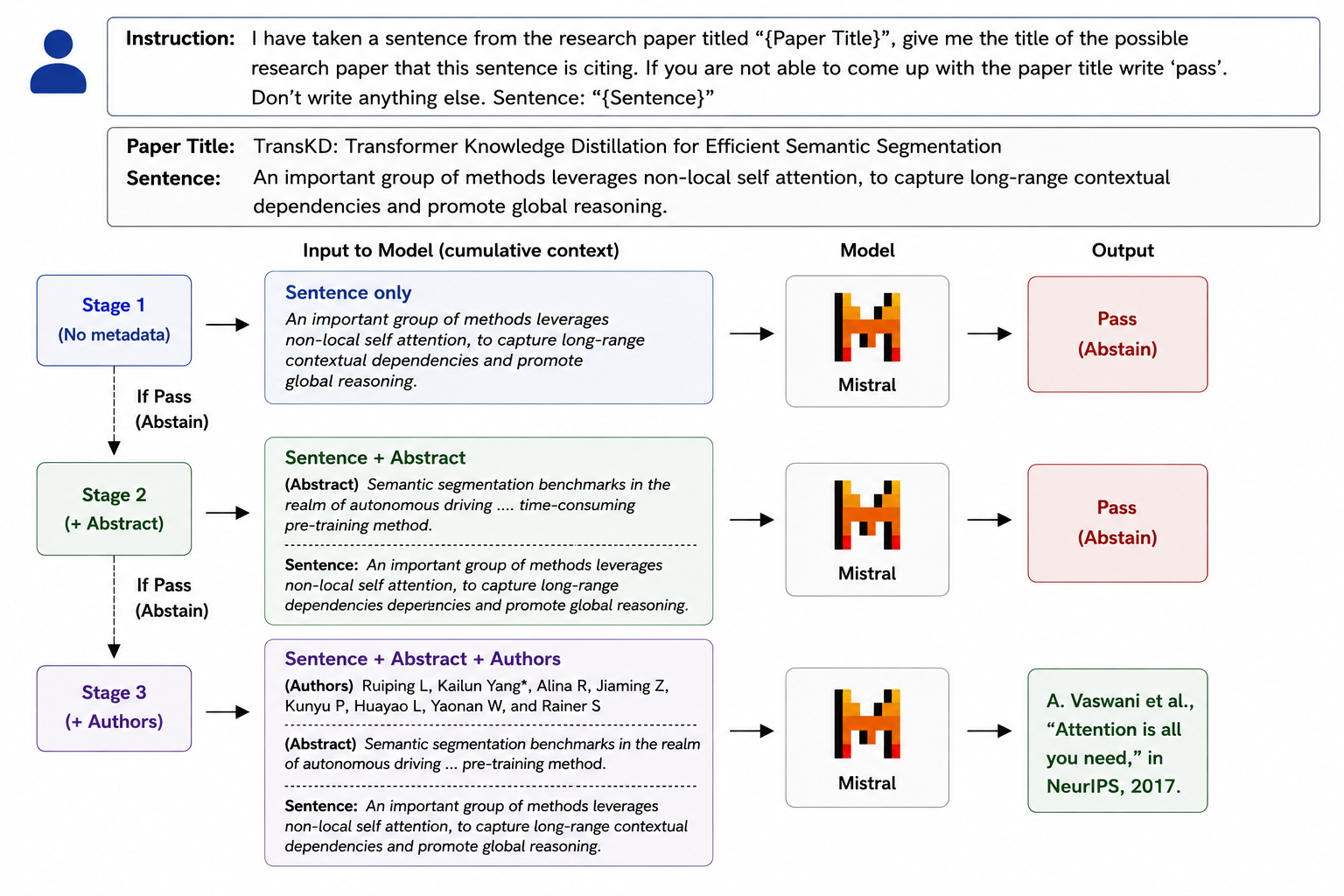}
  \captionsetup{aboveskip=2pt, belowskip=2pt} 
  \caption{\footnotesize CMP prompting as a staged attribution decision process. 
Under zero-shot indirect prompting (Task T2), the model abstains when asked to identify the cited paper from a sentence alone. 
As additional metadata is progressively revealed—first the abstract, then author information—the model transitions from abstention to confident attribution.}
  \label{fig:sid_exp}
\end{figure}

\paragraph{\textbf{Prompting Strategies}}We examine three prompting strategies. Let $x$ denote the query input (paper title for T1, citing sentence for T2), $m$ denote citation metadata (abstract, authors), and $f(\cdot)$ the model's generation function producing either an attribution $\hat{c}$ or the abstention token \texttt{Abstain (PASS)}. In \textit{zero-context prompting}, models receive only the query ($\hat{c} = f(x)$) and must either rely on parametric knowledge or abstain, thereby exposing baseline attribution behavior without retrieval. In \textit{metadata-augmented prompting}, models receive verified citation metadata alongside the query ($\hat{c} = f(x, m)$), revealing how structured evidence affects the abstention-hallucination tradeoff. In \textit{Cascaded metadata-augmented prompting (\textbf{CMP})}, the model first attempts T2 with no context; if it abstains or produces an error, we then provide metadata and re-query, enabling analysis of how models revise attribution decisions as evidence accumulates. This staged disclosure tests whether additional context shifts systems from cautious abstention to confident error. Figure~\ref{fig:sid_exp} illustrates CMP as a staged attribution decision process. CMP is not proposed as a training or calibration method but as a diagnostic tool: by staging evidence disclosure, it isolates whether a model's abstention behavior is driven by genuine uncertainty or simply by the absence of context. The key finding is that models with poor uncertainty calibration cannot be rescued by better prompting alone, motivating the need for calibration-aware training as future work.

\paragraph{\textbf{Retrieval Strategies}} Let $\mathcal{D}$ denote the retrieval corpus from \textsc{REASONS}. We evaluate two retrieval conditions that differ in how supporting documents are selected and provided to the model. $x$ is the query input, $f(\cdot)$ the model's generation
function, and $\hat{c}$ either an attribution or the abstention token
\textsc{Abstain}.


\noindent \textit{Naive retrieval augmentation.} A bi-encoder maps queries and documents into a shared embedding space. Documents are retrieved based on cosine similarity: $d_i \in \text{Top-K}_{d \in \mathcal{D}} \langle E(x), E(d) \rangle$, where $E(\cdot)$ denotes the embedding function and $\langle \cdot, \cdot \rangle$ denotes dot-product similarity. The top-$k$ retrieved documents are concatenated with the query: $\hat{c} = f(x, d_1 ..d_k)$. This represents standard dense retrieval without re-ranking.

\noindent \textit{Advanced retrieval augmentation.} The bi-encoder first retrieves a larger candidate set, which is then re-ranked using a cross-encoder that jointly scores each query-document pair. Documents are reordered according to $\pi(d \mid x) = \text{rank}(g(x, d))$, where $g(\cdot, \cdot)$ is the cross-encoder scoring function. The top-ranked documents from this reordered set are supplied to the generator: $\hat{c} = f(x, d_{\pi(1)}.. d_{\pi(k)})$. The cross-encoder is fine-tuned on \textsc{REASONS} using listwise context loss to optimize re-ranking. All retrieval conditions index the same corpus, $\mathcal{D}$, to ensure a controlled comparison.

\paragraph{\textbf{Models}}
We evaluate seven LLMs: GPT-o1 (G1), GPT-4o (G2), GPT-3.5-Turbo
(G3), pplx-7b-chat (P), DeepSeek-R1 (D),
Mistral-7B-v0.2-Instruct (M), and Llama-3.1-8B-Chat (L).
All models are evaluated on T1 and T2 under zero-context,
metadata-augmented, and CMP conditions, as applicable. For
retrieval-augmented evaluation, we use Mistral and Llama as
generators with Na\"ive RAG (RM, RL) and Advanced RAG (AM, AL).
Na\"ive RAG retrieves the top-$k=2$ documents, while Advanced
RAG retrieves $k=40$ candidates and re-ranks them using a
cross-encoder. Complete model and retrieval configurations are
provided in the Supplementary Material.

\noindent\textit{Citation-aware systems:}
We additionally evaluate GraphRAG~\citep{han2025retrieval},
HLM-Cite~\citep{hao2024hlm}, and LitFM~\citep{zhang2025litfm}
on T2. We provide the citing sentence as input and compare the
top-ranked citation against ground truth; incorrect predictions
are counted as hallucinations and null outputs as abstentions.
These systems are not evaluated on T1 because they perform
claim-to-citation mapping rather than metadata lookup. Results
are reported separately in Table~\ref{tab:retrieval_metrics};
implementation details are provided 
in the Supplementary Material, with retrieval components summarized
in Supplementary Table~\ref{imp_det}.

\paragraph{\textbf{Evaluation Metrics}}
We evaluate attribution behavior using metrics that distinguish abstention, incorrect attribution (hallucination), and partial surface overlap, rather than reducing performance to accuracy alone. 

\noindent \textit{Abstention Rate (AR).}
AR measures abstention frequency:
\[
\mathrm{AR} = \frac{1}{|\mathcal{Q}|} \sum_{i \in \mathcal{Q}} \mathbb{I}[\hat{c}_i = \textsc{pass}].
\]
where $\mathcal{Q}$ denotes the set of evaluation queries. 
AR reflects epistemic caution, not correctness. Abstention lacks instance-level ground truth—whether models should abstain depends on evidence availability. We validate AR through aggregate trends: it decreases with verified metadata but increases under adversarial perturbations (Supplementary Table ~\ref{tab:adv_metadata_t2}), demonstrating sensitivity to evidence quality rather than prompt compliance.

\noindent \textit{Hallucination Rate (HR).}
HR measures incorrect attribution frequency conditional on responding. Here, hallucination denotes any non-abstaining attribution that does
not match the benchmark ground truth, including both fabricated
citations and incorrect attribution to real sources. Let $\mathcal{Q}_{\neg \textsc{pass}} = \{ i \in \mathcal{Q} \mid \hat{c}_i \neq \textsc{pass} \}$:
\[
\mathrm{HR} = \frac{1}{|\mathcal{Q}_{\neg \textsc{pass}}|} 
\sum_{i \in \mathcal{Q}_{\neg \textsc{pass}}} 
\mathbb{I}[\hat{c}_i \neq c_i].
\]
Citation correctness is determined through exact string matching after standard text normalization. For title attribution (T2), the generated title must match the ground truth exactly. For author attribution (T1), we require the generated list to contain precisely the same authors as ground truth. Author matching is order-invariant and handles standard name variations (e.g., A. Smith $\equiv$ Andrew Smith). We treat partial matches as incorrect; incomplete author lists or truncated titles count as attribution failures. This strict criterion ensures HR measures identity-level precision rather than approximate similarity.

\noindent \textit{Joint interpretation.} AR and HR must be interpreted together to characterize epistemic safety. A system with high AR and low HR is cautious, it frequently declines to answer but rarely hallucinates when it does respond. Conversely, a system with low AR and high HR is overconfident; it answers most queries but produces many incorrect attributions. To illustrate the extremes: a model that always abstains achieves AR = 100\% with zero hallucinations but provides no utility, while a random citation from our corpus yields HR $\approx$ 99.99\% (only 1 in 12,723 citations would be correct by chance). Effective attribution systems must balance informativeness (low AR, high frequency of answers) with reliability (low HR, high accuracy). BLEU and F1 supplement HR to distinguish near-miss from spurious errors; correctness is validated through human evaluation on 1,000 stratified outputs.

\noindent\textit{Evaluation Process:}
We report performance aggregated across the 12 REASONS domains.
Domain-level variability is reported using standard deviations in
the Supplementary Material. Retrieval indices and re-ranking components remain fixed across all generator models, ensuring that observed differences reflect attribution behavior rather than variations in retrieval quality. Complete generation parameters,
prompt templates, and human validation protocols are
provided in the Supplementary Material.


\section{Results}

\tikzexternaldisable
\begin{figure*}[t]
\centering
\setlength{\abovecaptionskip}{2pt}
\setlength{\belowcaptionskip}{0pt}

\begin{subfigure}[b]{0.49\linewidth}\centering
\adjustbox{max width=\linewidth}{
\begin{tikzpicture}
    \begin{groupplot}[group style={group size=2 by 2,},width=4.5cm,height=3cm, xticklabel style={rotate=45, anchor=north east}]
     
     \nextgroupplot[
            symbolic x coords={G1, G2, G3, P, D, RM, M, RL, L, AL, AM},
            major tick length=0cm,
            xtick=data,
            ymin=0.0, ymax=0.8,
            enlarge x limits=0.1,
            enlarge y limits={upper,value=0.2},
            ybar,
            xticklabel style={font=\tiny},
            bar width = 4pt,
            title = {F-1 Score},
            title style={yshift=-2mm}
        ]
        \addplot+[
                error bars/.cd,
                y dir=both,
                y explicit
            ]    
        coordinates {
                (G1,0.40)  +- (0, 0.07)
                (G2, 0.424) +- (0, 0.09)
                (G3, 0.22) +- (0, 0.055)
                (P, 0.053) +- (0, 0.011)
                (D, 0.077) +- (0,0.007)
                (RM, 0.354) +- (0, 0.09)
                (M, 0.067) +- (0, 0.012)
                (RL, 0.253) +- (0, 0.069)
                (L, 0.349) +- (0, 0.024)
                (AL, 0.459) +- (0, 0.11)
                (AM, 0.574) +- (0, 0.07)
            };

     \nextgroupplot[
            symbolic x coords={G1, G2, G3, P, D, RM, M, RL, L, AL, AM},
            major tick length=0cm,
            xtick=data,
            ymin=0.0, ymax=0.8,
            enlarge x limits=0.1,
            enlarge y limits={upper,value=0.2},
            ybar,
            xticklabel style={font=\tiny},
            bar width = 4pt,
            title = {BLEU Score},
            title style={yshift=-2mm}
        ]
        \addplot+[
                error bars/.cd,
                y dir=both,
                y explicit
            ]    
        coordinates {
                (G1,0.358)  +- (0, 0.079)
                (G2, 0.337) +- (0, 0.098)
                (G3, 0.121) +- (0, 0.053)
                (P, 0.000) +- (0, 0.0)
                (D, 0.001) +- (0,0)
                (RM, 0.205) +- (0, 0.072)
                (M, 0.002) +- (0, 0.005)
                (RL, 0.134) +- (0, 0.048)
                (L,0.143) +- (0, 0.017)
                (AL, 0.421) +- (0, 0.160)
                (AM, 0.498) +- (0, 0.093)
            };

     \nextgroupplot[
            symbolic x coords={G1, G2, G3, P, D, RM, M, RL, L, AL, AM},
            major tick length=0cm,
            xtick=data,
            ymin=0.0, ymax=120,
            enlarge x limits=0.1,
            enlarge y limits={upper,value=0.2},
            ybar,
            xticklabel style={font=\tiny},
            bar width = 4pt,
            title = {Hallucination Rate},
            title style={yshift=-2mm}
        ]
        \addplot+[
                error bars/.cd,
                y dir=both,
                y explicit
            ]    
        coordinates {
                (G1,32.34)  +- (0, 8.52)
                (G2, 51.55) +- (0, 9.02)
                (G3, 71.73) +- (0, 5.80)
                (P, 94.67) +- (0, 1.58)
                (D, 92.05) +- (0,0)
                (RM, 62.53) +- (0,10.76)
                (M, 88.85) +- (0,14.25)
                (RL, 79.04) +- (0, 6.14)
                (L, 77.44) +- (0,1.73)
                (AL, 51.95) +- (0, 11.66)
                (AM, 43.34) +- (0, 7.75)
            };

     \nextgroupplot[
            symbolic x coords={G1, G2, G3, P, D, RM, M, RL, L, AL, AM},
            major tick length=0cm,
            xtick=data,
            ymin=0.0, ymax=110,
            enlarge x limits=0.1,
            enlarge y limits={upper,value=0.2},
            ybar,
            xticklabel style={font=\tiny},
            bar width = 4pt,
            title = {Abstention Rate},
            title style={yshift=-2mm}
        ]
        \addplot+[
                error bars/.cd,
                y dir=both,
                y explicit
            ]    
        coordinates {
                (G1,37.85)  +- (0, 8.069)
                (G2, 8.66) +- (0,3.502)
                (G3, 12.97) +- (0, 4.612)
                (P, 0.34) +- (0, 0.35)
                (D, 0.11) +- (0,0.09)
                (RM, 1.755) +- (0, 1.257)
                (M, 0.013) +- (0, 0.031)
                (RL, 0.0) +- (0, 0.0)
                (L, 0.0) +- (0, 0.0)
                (AL, 0) +- (0, 0)
                (AM, 0) +- (0, 0)
            };

    \end{groupplot}
\end{tikzpicture}
}
\caption{T1 zero-context.}\label{fig:T1_zero_compact}
\end{subfigure}\hfill
\begin{subfigure}[b]{0.49\linewidth}\centering
\adjustbox{max width=\linewidth}{\begin{tikzpicture}
    \begin{groupplot}[group style={group size=2 by 2,},width=4.5cm,height=3cm, xticklabel style={rotate=45, anchor=north east}]
     
     \nextgroupplot[
            symbolic x coords={G1, G2, G3, P, D, RM, M, RL, L, AL, AM},
            major tick length=0cm,
            xtick=data,
            ymin=0.0, ymax=1.0,
            enlarge x limits=0.1,
            enlarge y limits={upper,value=0.2},
            ybar,
            xticklabel style={font=\tiny},
            bar width = 4pt,
            title = {F-1 Score},
            title style={yshift=-2mm}
        ]
        \addplot+[
                error bars/.cd,
                y dir=both,
                y explicit
            ]    
        coordinates {
                (G1,0.9825)  +- (0, 0.0176)
                (G2,0.988) +- (0, 0.0057)
                (G3,0.955) +- (0, 0.006)
                (P, 0.500) +- (0, 0.112)
                (D, 0.044) +- (0,0.016)
                (RM, 0.567) +- (0, 0.167)
                (M,0.361) +- (0, 0.097)
                (RL, 0.359) +- (0, 0.067)
                (L,0.323) +- (0, 0.02)
                 (AL, 0.894) +- (0, 0.055)
                (AM, 0.884) +- (0, 0.158)
            };

     \nextgroupplot[
            symbolic x coords={G1, G2, G3, P, D, RM, M, RL, L, AL, AM},
            major tick length=0cm,
            xtick=data,
            ymin=0.0, ymax=1.2,
            enlarge x limits=0.1,
            enlarge y limits={upper,value=0.2},
            ybar,
            xticklabel style={font=\tiny},
            bar width = 4pt,
            title = {BLEU Score},
            title style={yshift=-2mm}
        ]
        \addplot+[
                error bars/.cd,
                y dir=both,
                y explicit
            ]    
        coordinates {
                (G1, 0.978)  +- (0,0.0174)
                (G2, 0.986) +- (0,0.0065)
                (G3, 0.935) +- (0,0.005)
                (P, 0.321) +- (0, 0.039)
                (D, 0.001) +- (0,0.001)
                (RM, 0.399) +- (0, 0.13)
                (M, 0.217) +- (0,0.071)
                (RL, 0.133) +- (0, 0.029)
                (L, 0.120) +- (0,0.009)
                 (AL, 0.875) +- (0, 0.056)
                (AM, 0.774) +- (0, 0.320)
            };

     \nextgroupplot[
            symbolic x coords={G1, G2, G3, P, D, RM, M, RL, L, AL, AM},
            major tick length=0cm,
            xtick=data,
            ymin=0.0, ymax=100,
            enlarge x limits=0.1,
            enlarge y limits={upper,value=0.2},
            ybar,
            xticklabel style={font=\tiny},
            bar width = 4pt,
            title = {Hallucination Rate},
            title style={yshift=-2mm}
        ]
        \addplot+[
                error bars/.cd,
                y dir=both,
                y explicit
            ]    
        coordinates {
                (G1, 0.39)  +- (0, 0.13)
                (G2, 0.13) +- (0, 0.07)
                (G3, 5.46) +- (0, 0.44)
                (P, 62.72) +- (0, 3.76)
                (D, 95.87) +- (0,0.01)
                (RM, 44.99) +- (0, 15.89)
                (M, 71.45) +- (0, 1.79)
                (RL, 75.36) +- (0, 4.52)
                (L, 80.282) +- (0, 0.909)
                 (AL, 13.94) +- (0, 6.67)
                (AM, 10.70) +- (0, 15.018)
            };

     \nextgroupplot[
            symbolic x coords={G1, G2, G3, P, D, RM, M, RL, L, AL, AM},
            major tick length=0cm,
            xtick=data,
            ymin=0.0, ymax=1.0,
            enlarge x limits=0.1,
            enlarge y limits={upper,value=0.2},
            ybar,
            xticklabel style={font=\tiny},
            bar width = 4pt,
            title = {Abstention Rate},
            title style={yshift=-2mm}
        ]
        \addplot+[
                error bars/.cd,
                y dir=both,
                y explicit
            ]    
        coordinates {
                (G1, 0.000)  +- (0, 0.00)
                (G2, 0.00) +- (0, 0.000)
                (G3, 0.00) +- (0, 0.00)
                (P, 0.13) +- (0, 0.21)
                (D, 0.13) +- (0,0.15)
                (RM, 0.09) +- (0, 0.22)
                (M, 0.0) +- (0, 0.00)
                (RL, 0.0) +- (0, 0.00)
                (L, 0.0) +- (0, 0.0)
                 (AL, 0) +- (0, 0)
                (AM, 0) +- (0, 0)
            };
    \end{groupplot}
\end{tikzpicture}
}
\caption{T1 +metadata.}\label{fig:T1_meta_compact}
\end{subfigure}

\vspace{1mm}

\begin{subfigure}[b]{0.49\linewidth}\centering
\adjustbox{max width=\linewidth}{\begin{tikzpicture}
    \begin{groupplot}[group style={group size=2 by 2,},width=4.5cm,height=3cm, xticklabel style={rotate=45, anchor=north east}]
    
    \nextgroupplot[
        symbolic x coords={G1, G2, G3, P, D, RM, M, RL, L},
        major tick length=0cm,
        xtick=data,
        ymin=0.0, ymax=0.5,
        enlarge x limits=0.1,
        enlarge y limits={upper,value=0.2},
        ybar,
        xticklabel style={font=\scriptsize},
        bar width = 5pt,
        title = {F-1 Score},
            title style={yshift=-2mm}
    ]
    \addplot+[
        error bars/.cd,
        y dir=both,
        y explicit
    ]    
    coordinates {
        (G1, 0.028)  +- (0, 0.02)
        (G2, 0.230) +- (0, 0.04)
        (G3, 0.00) +- (0, 0.00)
        (P, 0.006) +- (0, 0.020)
        (D, 0.22) +- (0, 0.15)
        (RM, 0.071) +- (0, 0.018)
        (M, 0.043) +- (0, 0.019)
        (RL, 0.0425) +- (0, 0.0154)
        (L, 0.029) +- (0, 0.015)
    };

    \nextgroupplot[
        symbolic x coords={G1, G2, G3, P, D, RM, M, RL, L},
        major tick length=0cm,
        xtick=data,
        ymin=0.0, ymax=0.5,
        enlarge x limits=0.1,
        enlarge y limits={upper,value=0.2},
        ybar,
        xticklabel style={font=\scriptsize},
        bar width = 5pt,
        title = {BLEU Score},
            title style={yshift=-2mm}
    ]
    \addplot+[
        error bars/.cd,
        y dir=both,
        y explicit
    ]    
    coordinates {
        (G1,0.019)  +- (0, 0.015)
        (G2, 0.100) +- (0, 0.038)
        (G3, 0.000) +- (0, 0.000)
        (P, 0.00) +- (0, 0.00)
        (D, 0.092) +- (0, 0.04)
        (RM, 0.038) +- (0, 0.026)
        (M, 0.00) +- (0, 0.00)
        (RL, 0.021) +- (0, 0.020)
        (L, 0.00) +- (0, 0.00)
    };

    \nextgroupplot[
        symbolic x coords={G1, G2, G3, P, D, RM, M, RL, L},
        major tick length=0cm,
        xtick=data,
        ymin=0.0, ymax=120,
        enlarge x limits=0.1,
        enlarge y limits={upper,value=0.1},
        ybar,
        xticklabel style={font=\scriptsize},
        bar width = 5pt,
        title = {Hallucination Rate},
            title style={yshift=-2mm}
    ]
    \addplot+[
        error bars/.cd,
        y dir=both,
        y explicit
    ]    
    coordinates {
        (G1, 67.73)  +- (0, 15.64)
        (G2, 72.49) +- (0, 5.51)
        (G3, 79.05) +- (0, 4.19)
        (P, 95.95) +- (0, 1.05)
        (D, 84.79) +- (0, 0.02)
        (RM, 93.389) +- (0, 1.40)
        (M, 96.04) +- (0, 3.45)
        (RL, 95.64) +- (0, 1.67)
        (L, 96.10) +- (0, 0.72)
    };

    \nextgroupplot[
        symbolic x coords={G1, G2, G3, P, D, RM, M, RL, L},
        major tick length=0cm,
        xtick=data,
        ymin=0.0, ymax=110,
        enlarge x limits=0.1,
        enlarge y limits={upper,value=0.2},
        ybar,
        xticklabel style={font=\scriptsize},
        bar width = 5pt,
        title = {Abstention Rate},
            title style={yshift=-2mm}
    ]
    \addplot+[
        error bars/.cd,
        y dir=both,
        y explicit
    ]    
    coordinates {
        (G1, 89.26)  +- (0, 5.52)
        (G2, 21.34) +- (0, 3.91)
        (G3, 97.50) +- (0, 0.94)
        (P, 89.17) +- (0, 28.11)
        (D, 16.77) +- (0, 2.61)
        (RM, 4.82) +- (0, 2.10)
        (M, 0.011) +- (0, 0.02)
        (RL, 0.0) +- (0, 0.0)
        (L, 0.0) +- (0, 0.0)
    };

    \end{groupplot}
\end{tikzpicture}
}
\caption{T2 zero-context.}\label{fig:T2_zero_compact}
\end{subfigure}\hfill
\begin{subfigure}[b]{0.49\linewidth}\centering
\adjustbox{max width=\linewidth}{\begin{tikzpicture}
    \begin{groupplot}[group style={group size=2 by 2,},width=4.5cm,height=3cm, xticklabel style={rotate=45, anchor=north east}]
     
     \nextgroupplot[
            symbolic x coords={G1, G2, G3, P, D, RM, M, RL, L, AL, AM},
            major tick length=0cm,
            xtick=data,
            ymin=0.0, ymax=0.8,
            enlarge x limits=0.1,
            enlarge y limits={upper,value=0.2},
            ybar,
            xticklabel style={font=\tiny},
            bar width = 4pt,
            title = {F-1 Score},
            title style={yshift=-2mm}
        ]
        \addplot+[
                error bars/.cd,
                y dir=both,
                y explicit
            ]    
        coordinates {
                (G1, 0.281)  +- (0, 0.094)
                (G2, 0.468) +- (0, 0.224)
                (G3, 0.043) +- (0, 0.017)
                (P, 0.073) +- (0, 0.014)
                (D, 0.192) +- (0,0.028)
                (RM, 0.124) +- (0, 0.028)
                (M, 0.094) +- (0, 0.026)
                (RL, 0.156) +- (0, 0.041)
                (L, 0.050) +- (0, 0.11)
                (AL, 0.498) +- (0, 0.141)
                (AM, 0.588) +- (0, 0.147)
            };

     \nextgroupplot[
            symbolic x coords={G1, G2, G3, P, D, RM, M, RL, L, AL, AM},
            major tick length=0cm,
            xtick=data,
            ymin=0.0, ymax=0.7,
            enlarge x limits=0.1,
            enlarge y limits={upper,value=0.2},
            ybar,
            xticklabel style={font=\tiny},
            bar width = 4pt,
            title = {BLEU Score},
            title style={yshift=-2mm}
        ]
        \addplot+[
                error bars/.cd,
                y dir=both,
                y explicit
            ]    
        coordinates {
                (G1, 0.228)  +- (0, 0.079)
                (G2, 0.284) +- (0, 0.158)
                (G3, 0.018) +- (0, 0.011)
                (P, 0.006) +- (0, 0.009)
                (D, 0.03) +- (0,0)
                (RM, 0.046) +- (0, 0.020)
                (M, 0.003) +- (0, 0.004)
                (RL, 0.039) +- (0, 0.020)
                (L, 0.022) +- (0, 0.075)
                (AL, 0.347) +- (0, 0.142)
                (AM, 0.453) +- (0, 0.145)
            };

     \nextgroupplot[
            symbolic x coords={G1, G2, G3, P, D, RM, M, RL, L, AL, AM},
            major tick length=0cm,
            xtick=data,
            ymin=0.0, ymax=120,
            enlarge x limits=0.1,
            enlarge y limits={upper,value=0.2},
            ybar,
            xticklabel style={font=\tiny},
            bar width = 4pt,
            title = {Hallucination Rate},
            title style={yshift=-2mm}
        ]
        \addplot+[
                error bars/.cd,
                y dir=both,
                y explicit
            ]    
        coordinates {
                (G1, 33.65)  +- (0, 7.80)
                (G2, 51.42) +- (0, 8.85)
                (G3, 64.49) +- (0, 7.16)
                (P, 94.40) +- (0, 1.51)
                (D, 84.04) +- (0,0.02)
                (RM, 87.87) +- (0, 3.25)
                (M, 93.65) +- (0, 2.38)
                (RL, 89.10) +- (0, 2.85)
                (L, 95.37) +- (0, 11.51)
                (AL, 53.78) +- (0, 10.60)
                (AM, 43.04) +- (0, 10.84)
            };

     \nextgroupplot[
            symbolic x coords={G1, G2, G3, P, D, RM, M, RL, L, AL, AM},
            major tick length=0cm,
            xtick=data,
            ymin=0.0, ymax=110,
            enlarge x limits=0.1,
            enlarge y limits={upper,value=0.2},
            ybar,
            xticklabel style={font=\tiny},
            bar width = 4pt,
            title = {Abstention Rate},
            title style={yshift=-2mm}
        ]
        \addplot+[
                error bars/.cd,
                y dir=both,
                y explicit
            ]    
        coordinates {
                (G1, 59.05)  +- (0, 9.92)
                (G2, 18.331) +- (0, 33.53)
                (G3, 87.31) +- (0, 2.63)
                (P, 1.30) +- (0, 1.02)
                (D, 0.93) +- (0,0.52)
                (RM, 5.3575) +- (0, 1.971)
                (M, 0.0) +- (0, 0.0)
                (RL, 0.065) +- (0, 0.22)
                (L, 0.0) +- (0, 0.0)
                (AL, 0) +- (0, 0.)
                (AM, 0) +- (0, 0)
            };

    \end{groupplot}
\end{tikzpicture}
}
\caption{T2 staged (CMP).}\label{fig:T2_sid_compact}
\end{subfigure}

\vspace{1mm}
\caption{\footnotesize Attribution behavior (AR/HR) across models and tasks, averaged over 12 REASONS domains. RAG-based systems (\texttt{RM}, \texttt{RL}, \texttt{AL}, \texttt{AM}) suppress abstention even when hallucination remains high, while \texttt{G1} and \texttt{G2} retain higher AR at the cost of responsiveness. Other models (\texttt{G3}, \texttt{M}, \texttt{L}, \texttt{D}, \texttt{P}) tend toward overconfident hallucination, revealing distinct epistemic risk profiles across model and retrieval configurations. Error bars show variability across the 12 domains.}
\label{fig:compact_all}
\end{figure*}
\tikzexternalenable


Figure~\ref{fig:compact_all} reports AR (refusal rate) and HR (error rate when answering), with error bars showing variability across 12 domains, revealing model caution versus overconfidence.

\textit{\textbf{Baseline attribution behavior under minimal context:}} Under zero-context prompting, attribution is unreliable for both tasks, but failures differ by task type.
For T1 (direct attribution), proprietary models achieve moderate AR ($\equiv$35–40\%) yet still exhibit substantial HR ($>$30\%) among non-abstaining outputs (Figure~\ref{fig:T1_zero_compact}). Smaller or task-specific systems show extremely high HR (often $>$90\%), indicating strong ``authority-assertion” tendencies even when evidence is absent. T2 (indirect attribution) is consistently harder: HR increases sharply across models (often $>$70\%), and model strategies diverge. GPT-4–class models respond by abstaining heavily (AR approaching ~90\%), while RAG/task-specific systems continue answering despite high error (Figure~\ref{fig:T2_zero_compact}). Zero-context diagnoses whether a model defaults to calibrated caution (high AR, low HR) or overconfident fabrication (low AR, high HR).

\begin{table}[t]
\centering
\setlength{\tabcolsep}{4pt}
\renewcommand{\arraystretch}{1.18}
\resizebox{\columnwidth}{!}{%
\begin{tabular}{@{}l l c c c@{}}
\toprule
\textbf{Category} & \textbf{Domain} & \textbf{\# Papers} & \textbf{HR (Zero-Context)} & \textbf{HR (Metadata)} \\
\midrule

\multirow{2}{*}[-0.4ex]{Well-represented}
& CV & 5488 & 71.3\% & 12.8\% \\
& IR & 1741 & 68.9\% & 14.2\% \\

\midrule

\multirow{2}{*}[-0.4ex]{Specialized}
& QC  & 421 & 94.6\% & 31.7\% \\
& Bio & 119 & 96.1\% & 28.9\% \\

\bottomrule
\end{tabular}}
\caption{\footnotesize Hallucination Rate (HR) stratified by domain representation.
Specialized, low-resource domains exhibit substantially higher epistemic risk than well-represented domains,
even when metadata grounding is provided.}
\label{tab:domain_hr}
\end{table}

\noindent \textit{\textbf{Effect of metadata on attribution behavior:}} 
Adding verified metadata drastically reduces HR for T1, and AR collapses; models answer because the evidence is explicit (Figure~\ref{fig:T1_meta_compact}). Metadata reduces error, but can also shift the failure mode. For T2, HR drops but remains nontrivial, while AR often approaches zero for many models despite residual mistakes. T1 becomes almost “copy-from-evidence.” Authorship is explicitly present in metadata, so the task is closer to extraction than inference. T2 still requires alignment and selection. Even with metadata, models must connect a cited claim to the correct paper/title among plausible candidates; this invites surface similarity bias (choosing what “looks right” rather than what is correct). Metadata can create evidence-induced confidence: models stop abstaining because something looks grounded, but the grounding may be misapplied (wrong paper/title is selected), leaving HR at 15–20\% for some models.

\noindent \textit{\textbf{RAG models:}} 
RAG variants reduce HR relative to zero-context baselines and improve F1/BLEU, but they often show near-zero AR even when HR remains high (Table~\ref{tab:retrieval_metrics}). Naive RAG is especially prone to confident errors; advanced RAG reduces HR but does not restore abstention. Retrieval injects text that looks like evidence, which increases the model’s tendency to answer. Therefore, retrieval strength is not a proxy for reliability; AR–HR jointly identify whether RAG is helping correctness or merely suppressing abstention.

\noindent \textit{\textbf{CMP}} reduces HR and improves attribution quality relative to zero-context, but model families respond differently: GPT-4–class models retain higher AR, while RAG-based systems still suppress AR and keep asserting attributions (Figure~\ref{fig:T2_sid_compact}). Staging prevents models from committing prematurely by forcing incremental reasoning. If a model always answers when given context, adding more stages does not help it abstain. CMP works best for models that already know when to say
``I don't know.'' For overconfident models, it reduces errors
somewhat but does not fix the abstention problem.

\begin{table}[t]
\centering
\small
\begin{tabular}{@{}lcccc@{}}
\toprule
\textbf{Method} & \textbf{AR (\%)} & \textbf{HR (\%)} & \textbf{F1} & \textbf{BLEU} \\
\midrule
Naive RAG  & 5.03 & 87.56 & 0.14 & 0.09 \\
Advanced RAG  & \textbf{0.00} & \textbf{65.41} & \textbf{0.58} & \textbf{0.32} \\
GraphRAG & 2.14 & 68.33 & 0.52 & 0.38 \\
HLM-Cite & 12.45 & 71.22 & 0.41 & 0.31 \\
LitFM & 8.91 & 74.16 & 0.48 & 0.35 \\
\bottomrule
\end{tabular}
\caption{\footnotesize Citation-aware systems trade abstention for responsiveness.
GraphRAG exhibits near-zero AR, while HLM-Cite retains limited
abstention. F1 and BLEU reveal that even lower-HR systems produce
only partial matches, indicating that attribution errors are systematic
rather than incidental.}
\label{tab:retrieval_metrics}
\end{table}

\noindent \textit{\textbf{Domain effects across tasks (Specialized vs. Well-represented):}} Specialized domains (QC, Biomolecules) show consistently higher HR than well-represented ones (CV, IR), even with metadata (Table~\ref{tab:domain_hr}). Subsampling to equalize domain size preserves the gap, indicating the effect is not just data volume. Specialized fields use dense jargon, uncommon venues, and long-tail author/title distributions, conditions where “plausible guessing” is easier than correct identification. Retrieval is also noisier: similar terms or nearby topics produce highly plausible but wrong candidates. Domain specialization is a structural risk factor: even good prompting and retrieval can leave substantial residual HR, so deployment in specialized areas needs stronger verification than “more context.”

\noindent \textit{\textbf{Summary:}} We observe three recurring behaviors: (1) cautious abstention (high AR, low HR); (2) confident hallucination (low AR, high HR); and (3) metadata-induced confidence, where added context suppresses abstention without eliminating error. These regimes emerge from interactions among task type, evidence availability, retrieval design, and domain structure.

\noindent \textit{\textbf{Human validation of HR Metric:}} To validate that HR accurately captures attribution behavior, we conduct a human evaluation on a stratified sample of \textbf{1,000} model outputs. These instances were stratified across three dimensions to ensure
representativeness: (a) \textit{Model coverage}: 200 instances each from GPT-o1, GPT-4o, Advanced RAG
(Mistral), Perplexity AI (7B), and DeepSeek; (b) \textit{Domain diversity}: Proportional sampling across well-represented domains (CV, IR) and specialized domains (QC, Biomolecules); (c) \textit{Behavioral distribution}: Deliberate oversampling of abstention cases yields 500 non-pass instances (50\%) and 500 pass instances (50\%), enabling robust analysis of both attribution correctness and abstention appropriateness. This balanced distribution contrasts with the natural model behavior (which exhibits $>$90\% non-pass rates) and provides sufficient power to analyze abstention decisions, a critical but underrepresented behavior in typical model outputs.

\noindent \textit{Annotation protocol.} Three domain experts (PhD students with $>$3 years of research experience) independently evaluate each instance. Annotators receive: (1) the input query (paper title for T1, cited sentence for T2); (2) the model output (attribution or \texttt{PASS}); (3) ground-truth citation metadata (title, authors, abstract); (4) evidence condition (zero-context, metadata-augmented, or CMP).

Annotators apply a two-stage judgment process depending on model behavior: \textit{Stage 1. For non-pass outputs (n=500):} Rate attribution correctness:
(a) Correct: Model output matches ground truth semantically, allowing minor formatting variations (e.g., ``Reliable Attribution in Scientific Language Models'' vs. ``A Trustworthy Attribution for Scholarly Language Models''); (b) Partially Correct: Model identifies the correct paper but with incomplete information (e.g., missing co-authors, abbreviated title); (c) Incorrect: Model output does not match ground truth (hallucination). 

\noindent \textit{Stage 2---Pass outputs (n=500):} Annotators judged whether each abstention was appropriate given the available evidence. An abstention was marked appropriate if the model correctly declined when evidence was insufficient (e.g., zero-context with an ambiguous query), and unnecessary if the model declined despite having sufficient evidence (e.g., complete metadata provided).

\noindent \textit{Inter-annotator agreement.} Fleiss' $\kappa$ was computed separately for attribution correctness ($\kappa = 0.81$) and abstention appropriateness ($\kappa = 0.74$), yielding an overall $\kappa = 0.78$, indicating substantial agreement across all judgment types. 
Disagreements were resolved by majority vote. Among the 500
non-pass instances, 330 (66.0\%) were judged factually incorrect.
Among the 500 pass instances, 271 (54.2\%) abstentions were judged
appropriate and 229 (45.8\%) unnecessary.

\noindent \textit{Validation of automated metrics.} We validate automated HR by comparing it to human-judged error rates on the same outputs (Table~\ref{tab:human_validation}), confirming strong alignment despite minor paraphrase variation. The strong correlation ($r = 0.89$, $p < 0.001$) between automated HR and human-judged error rates demonstrates that:
1) Lexical-level metrics effectively capture attribution errors without being confounded by acceptable paraphrases. 2) When models hallucinate, they typically produce factually different content (wrong paper, wrong authors), not mere rewordings. 3) Small divergences (e.g., G1: 17.99\% automated vs. 19.39\% human) reflect cases where human experts identified subtle errors (author name typos, venue misattributions) that lexical metrics missed.

\begin{table}[t!]
\centering
\footnotesize
\setlength{\tabcolsep}{3.5pt}
\renewcommand{\arraystretch}{1.15}
\begin{tabular}{l c c c c}
\toprule
\textbf{Model} & \textbf{n} & \textbf{Auto HR} & \textbf{Human Incorrect} & \textbf{r} \\
 & \textbf{(non-pass)} & \textbf{(\%)} & \textbf{(\%)} &  \\
\midrule
G1 & 98  & 17.99 & 19.39 & 0.91 \\
G2 & 97  & 17.59 & 18.39 & 0.85 \\
AM & 102 & 75.41 & 73.53 & 0.88 \\
P  & 101 & 86.78 & 84.16 & 0.87 \\
D  & 102 & 71.34 & 69.61 & 0.90 \\
\midrule
\textbf{Total} & \textbf{500} & \textbf{--} & \textbf{--} & \textbf{0.89} \\
\bottomrule
\end{tabular}
\caption{\footnotesize Human evaluation validates automated HR measurement. HR is computed only over non-pass instances and compared against human judgments of attribution correctness on the same subset. Strong correlation ($r = 0.89$, $p < 0.001$) confirms that lexical-level metrics capture human-perceived errors with minimal confounding from acceptable paraphrases.}
\label{tab:human_validation}
\end{table}

Only 5.2\% of non-pass outputs corresponded to acceptable paraphrases, indicating that most detected errors were factual attribution failures rather than lexical variations. \textit{Additional analyses of paraphrase and abstention behavior are provided in the Supplementary Material.}

\section{Ethical considerations}

The dataset consists of publicly available arXiv papers; we follow
arXiv usage policies, exclude restricted content, and use no private
or personal data. This is an observational
study of system behavior, and we do not release fine-tuned generator
checkpoints trained on restricted material. We evaluate two tasks:
\textbf{T1} (direct attribution), which tests whether systems assert or
withhold authority cues, and \textbf{T2} (indirect attribution), which
tests whether systems link claims to sources. Across both tasks, we
characterize behavior via the joint dynamics of \textbf{AR} and
\textbf{HR}; unless stated otherwise, results are averaged over all 12
domains. Because \textsc{REASONS} is derived from arXiv, some benchmark
instances may overlap with LLM pretraining corpora. T1 may partly reflect
memorized title--author associations, whereas T2 requires claim-to-source
attribution and is therefore less directly dependent on metadata recall.

\textsc{REASONS} evaluates attribution fidelity to original author
citation choices, not attribution correctness in an absolute sense.
Authors may miscite or overcite, meaning a model that identifies a more
appropriate citation than the original may still be penalized. Author
citation choices provide scalable ground truth for this benchmark, and
the same evaluation criterion is applied consistently across models.
The human annotations are not released because of institutional review
requirements; however, we release the annotation rubric, protocol,
examples, and aggregation procedure to facilitate reproduction.

\section{Conclusion} We introduced \textsc{REASONS}, a sentence-level benchmark for scientific source attribution across 12 arXiv subject categories, and used it to systematically evaluate how LLMs attribute paper \emph{titles} and \emph{authors} under varying evidence conditions. Our experiments reveal a consistent \emph{abstention--hallucination} tradeoff: models that answer more readily often reduce apparent abstention but may increase confident misattributions, especially when contextual cues resemble authoritative metadata. RAG and enriched citation context typically improve attribution fidelity, yet can also suppress abstention even when the retrieved evidence is incomplete or misleading, creating a different form of reliability risk. Our results suggest that evaluating scientific attribution systems requires metrics beyond accuracy alone, explicitly accounting for when models should abstain. Our findings are limited to the 12 arXiv subject categories studied
and should not be assumed to generalize to other scientific disciplines. Future work includes improving evidence-grounding under noisy retrieval, designing calibration methods that preserve appropriate abstention, and extending the benchmark to broader citation styles and full-document attribution settings.

\bibliographystyle{IEEEtran}
\bibliography{aaai2026}

@article{lewis2020retrieval,
  title={Retrieval-augmented generation for knowledge-intensive nlp tasks},
  author={Lewis, Patrick and Perez, Ethan and Piktus, Aleksandra and Petroni, Fabio and Karpukhin, Vladimir and Goyal, Naman and K{\"u}ttler, Heinrich and Lewis, Mike and Yih, Wen-tau and Rockt{\"a}schel, Tim and others},
  journal={Advances in Neural Information Processing Systems},
  volume={33},
  pages={9459--9474},
  year={2020}
}

@article{tilwani2026neurosymbolic,
  title={Neurosymbolic AI for Legal AI-TRISM: Trustworthy, Reliable, Interpretable, Safe Models},
  author={Tilwani, Deepa and Saxena, Yash and Padia, Ankur and Parthasarathy, Srinivasan and Gaur, Manas},
  journal={Neurosymbolic AI: Foundations and Applications},
  pages={429--454},
  year={2026},
  publisher={Wiley Online Library}
}

@article{tilwani2024neurosymbolic,
  title={Neurosymbolic ai approach to attribution in large language models},
  author={Tilwani, Deepa and Venkataramanan, Revathy and Sheth, Amit P},
  journal={IEEE Intelligent Systems},
  volume={39},
  number={6},
  pages={10--17},
  year={2024},
  publisher={IEEE}
}

@inproceedings{khalifasource,
  title={Source-Aware Training Enables Knowledge Attribution in Language Models},
  author={Khalifa, Muhammad and Wadden, David and Strubell, Emma and Lee, Honglak and Wang, Lu and Beltagy, Iz and Peng, Hao},
  booktitle={First Conference on Language Modeling},
  year = {2024}
  
}

@inproceedings{saxena2025generation,
  title={Generation-Time vs. Post-hoc Citation: A Holistic Evaluation of LLM Attribution},
  author={Saxena, Yash and Bommireddy, Raviteja and Padia, Ankur and Gaur, Manas},
  booktitle={NeurIPS 2025 Workshop on Evaluating the Evolving LLM Lifecycle: Benchmarks, Emergent Abilities, and Scaling},
  year = {2025}
}

@inproceedings{gururangan2020don,
  title={Don’t Stop Pretraining: Adapt Language Models to Domains and Tasks},
  author={Gururangan, Suchin and Marasovi{\'c}, Ana and Swayamdipta, Swabha and Lo, Kyle and Beltagy, Iz and Downey, Doug and Smith, Noah A},
  booktitle={Proceedings of the 58th Annual Meeting of the Association for Computational Linguistics},
  pages={8342--8360},
  year={2020}
}

@online{mehdi2023reinventing,
  author       = {Mehdi, Yusuf},
  title        = {Reinventing Search with a New AI-Powered Microsoft Bing and Edge: Your Copilot for the Web},
  year         = {2023},
  month        = feb,
  day          = {7},
  url          = {https://blogs.microsoft.com/blog/2023/02/07/},
  note         = {Microsoft Official Blog},
}

@misc{roose2024can,
  author       = {Kevin Roose},
  title        = {Can This A.I.-Powered Search Engine Replace Google? It Has for Me},
  howpublished = {\emph{The New York Times}},
  year         = {2024},
  month        = feb,
  note         = {Published February 1, 2024}
}

@article{xu2024hallucination,
  title={Hallucination is inevitable: An innate limitation of large language models},
  author={Xu, Ziwei and Jain, Sanjay and Kankanhalli, Mohan},
  journal={arXiv preprint arXiv:2401.11817},
  year={2024}
}

@article{walters2023fabrication,
  title={Fabrication and errors in the bibliographic citations generated by ChatGPT},
  author={Walters, William H and Wilder, Esther Isabelle},
  journal={Scientific Reports},
  volume={13},
  number={1},
  pages={14045},
  year={2023},
  publisher={Nature Publishing Group UK London}
}

@online{ShmatkoAdamEsau2026NeurIPS,
  title        = {GPTZero finds 100 new hallucinations in NeurIPS 2025 accepted papers},
  author       = {Shmatko, Nazar and Adam, Alex and Esau, Paul},
  year         = {2026},
  month        = jan,
  day          = {21},
  organization = {GPTZero},
  url          = {https://gptzero.me/news/neurips/},
  urldate      = {2026-02-09}
}

@inproceedings{cohan2019structural,
  title={Structural Scaffolds for Citation Intent Classification in Scientific Publications},
  author={Cohan, Arman and Ammar, Waleed and van Zuylen, Madeleine and Cady, Field},
  booktitle={Proceedings of the 2019 Conference of the North American Chapter of the Association for Computational Linguistics: Human Language Technologies, Volume 1 (Long and Short Papers)},
  pages={3586--3596},
  year={2019},
  publisher={Association for Computational Linguistics},
  doi={10.18653/v1/N19-1361}
}

@inproceedings{lo2020s2orc,
  title={S2ORC: The Semantic Scholar Open Research Corpus},
  author={Lo, Kyle and Wang, Lucy Lu and Neumann, Mark and Kinney, Rodney and Weld, Daniel S},
  booktitle={Proceedings of the 58th Annual Meeting of the Association for Computational Linguistics},
  pages={4969--4983},
  year={2020},
  publisher={Association for Computational Linguistics},
  doi={10.18653/v1/2020.acl-main.447}
}

@article{saier2020unarxive,
  title={unarXive: a large scholarly data set with publications’ full-text, annotated in-text citations, and links to metadata},
  author={Saier, Tarek and F{\"a}rber, Michael},
  journal={Scientometrics},
  volume={125},
  number={3},
  pages={3085--3108},
  year={2020},
  publisher={Springer}
}

@inproceedings{liu-etal-2021-refsum,
    title = "{R}ef{S}um: Refactoring Neural Summarization",
    author = "Liu, Yixin  and
      Dou, Zi-Yi  and
      Liu, Pengfei",
    editor = "Toutanova, Kristina  and
      Rumshisky, Anna  and
      Zettlemoyer, Luke  and
      Hakkani-Tur, Dilek  and
      Beltagy, Iz  and
      Bethard, Steven  and
      Cotterell, Ryan  and
      Chakraborty, Tanmoy  and
      Zhou, Yichao",
    booktitle = "Proceedings of the 2021 Conference of the North American Chapter of the Association for Computational Linguistics: Human Language Technologies",
    month = jun,
    year = "2021",
    address = "Online",
    publisher = "Association for Computational Linguistics",
    url = "https://aclanthology.org/2021.naacl-main.113/",
    doi = "10.18653/v1/2021.naacl-main.113",
    pages = "1437--1448",
}

@article{hao2024hlm,
  title={Hlm-cite: Hybrid language model workflow for text-based scientific citation prediction},
  author={Hao, Qianyue and Fan, Jingyang and Xu, Fengli and Yuan, Jian and Li, Yong},
  journal={Advances in Neural Information Processing Systems},
  volume={37},
  pages={48189--48223},
  year={2024}
}

@inproceedings{zhang2025litfm,
  title={Litfm: A retrieval augmented structure-aware foundation model for citation graphs},
  author={Zhang, Jiasheng and Maatouk, Ali and Chen, Jialin and Bui, Ngoc and Xie, Qianqian and Tassiulas, Leandros and Xu, Hua and Shao, Jie and Ying, Rex},
  booktitle={Proceedings of the 31st ACM SIGKDD Conference on Knowledge Discovery and Data Mining V. 2},
  pages={3728--3739},
  year={2025}
}

@article{han2025retrieval,
  title={Retrieval-Augmented Generation with Graphs (GraphRAG)},
  author={Han, Haoyu and Wang, Yu and Shomer, Harry and Guo, Kai and Ding, Jiayuan and Lei, Yongjia and Halappanavar, Mahantesh and Rossi, Ryan A and Mukherjee, Subhabrata and Tang, Xianfeng and others},
  journal={CoRR},
  year={2025}
}

@misc{abolghasemi2024,
      title={Evaluation of Attribution Bias in Retrieval-Augmented Large Language Models}, 
      author={Amin Abolghasemi and Leif Azzopardi and Seyyed Hadi Hashemi and Maarten de Rijke and Suzan Verberne},
      year={2024},
      eprint={2410.12380},
      archivePrefix={arXiv},
      primaryClass={cs.CL},
      url={https://arxiv.org/abs/2410.12380}, 
}

@inproceedings{huang-etal-2024-advancing,
    title = "Advancing Large Language Model Attribution through Self-Improving",
    author = "Huang, Lei  and
      Feng, Xiaocheng  and
      Ma, Weitao  and
      Zhao, Liang  and
      Fan, Yuchun  and
      Zhong, Weihong  and
      Xu, Dongliang  and
      Yang, Qing  and
      Liu, Hongtao  and
      Qin, Bing",
    editor = "Al-Onaizan, Yaser  and
      Bansal, Mohit  and
      Chen, Yun-Nung",
    booktitle = "Proceedings of the 2024 Conference on Empirical Methods in Natural Language Processing",
    month = nov,
    year = "2024",
    address = "Miami, Florida, USA",
    publisher = "Association for Computational Linguistics",
    url = "https://aclanthology.org/2024.emnlp-main.223/",
    doi = "10.18653/v1/2024.emnlp-main.223",
    pages = "3822--3836",
}

@article{wagner2025,
author = {Wagner, Christian and Jiang, Ling},
title = {Death by AI: Will large language models diminish Wikipedia?},
journal = {Journal of the Association for Information Science and Technology},
volume = {76},
number = {5},
pages = {743-751},
year = {2025}
}

@article{bohnet2022attributed,
  title={Attributed question answering: Evaluation and modeling for attributed large language models},
  author={Bohnet, Bernd and Tran, Vinh Q and Verga, Pat and Aharoni, Roee and Andor, Daniel and Soares, Livio Baldini and Ciaramita, Massimiliano and Eisenstein, Jacob and Ganchev, Kuzman and Herzig, Jonathan and others},
  journal={arXiv preprint arXiv:2212.08037},
  year={2022}
}

@inproceedings{lin2022truthfulqa,
  title={{TruthfulQA}: Measuring How Models Mimic Human Falsehoods},
  author={Lin, Stephanie and Hilton, Jacob and Evans, Owain},
  booktitle={Proceedings of the 60th Annual Meeting of the Association for Computational Linguistics (Volume 1: Long Papers)},
  pages={3214--3252},
  year={2022},
  publisher={Association for Computational Linguistics},
  address={Dublin, Ireland},
  doi={10.18653/v1/2022.acl-long.229},
  url={https://aclanthology.org/2022.acl-long.229}
}

@InProceedings{chuang2025selfcite,
  title = 	 {{S}elf{C}ite: Self-Supervised Alignment for Context Attribution in Large Language Models},
  author =       {Chuang, Yung-Sung and Cohen-Wang, Benjamin and Shen, Zejiang and Wu, Zhaofeng and Xu, Hu and Lin, Xi Victoria and Glass, James R. and Li, Shang-Wen and Yih, Wen-Tau},
  booktitle = 	 {Proceedings of the 42nd International Conference on Machine Learning},
  pages = 	 {10839--10858},
  year = 	 {2025},
  editor = 	 {Singh, Aarti and Fazel, Maryam and Hsu, Daniel and Lacoste-Julien, Simon and Berkenkamp, Felix and Maharaj, Tegan and Wagstaff, Kiri and Zhu, Jerry},
  volume = 	 {267},
  series = 	 {Proceedings of Machine Learning Research},
  month = 	 {13--19 Jul},
  publisher =    {PMLR},
}

@inproceedings{borgeaud2022improving,
  title={Improving Language Models by Retrieving from Trillions of Tokens},
  author={Borgeaud, Sebastian and Mensch, Arthur and Hoffmann, Jordan and Cai, Trevor and Rutherford, Eliza and Millican, Katie and Van Den Driessche, George Brant and Lespiau, Jean-Baptiste and Damoc, Bogdan and Clark, Aidan and de Las Casas, Diego and Guy, Aurelia and Menick, Jacob and Ring, Roman and Hennigan, Tom and Huang, Saffron and Maggiore, Loren and Jones, Chris and Cassirer, Albin and Brock, Andy and Paganini, Michela and Irving, Geoffrey and Vinyals, Oriol and Osindero, Simon and Simonyan, Karen and Rae, Jack and Elsen, Erich and Sifre, Laurent},
  booktitle={Proceedings of the 39th International Conference on Machine Learning},
  pages={2206--2240},
  year={2022},
  volume={162},
  series={Proceedings of Machine Learning Research},
  publisher={PMLR},
  url={https://proceedings.mlr.press/v162/borgeaud22a.html}
}

@inproceedings{petroni2020kilt,
  title={{KILT}: A Benchmark for Knowledge Intensive Language Tasks},
  author={Petroni, Fabio and Piktus, Aleksandra and Fan, Angela and Lewis, Patrick and Yazdani, Majid and De Cao, Nicola and Thorne, James and Jernite, Yacine and Karpukhin, Vladimir and Maillard, Jean and Plachouras, Vassilis and Rockt{\"a}schel, Tim and Riedel, Sebastian},
  booktitle={Proceedings of the 2021 Conference of the North American Chapter of the Association for Computational Linguistics: Human Language Technologies},
  pages={2523--2544},
  year={2021},
  publisher={Association for Computational Linguistics},
  address={Online},
  doi={10.18653/v1/2021.naacl-main.200},
  url={https://aclanthology.org/2021.naacl-main.200}
}

@inproceedings{thorne-etal-2018-fever,
  title     = "{FEVER}: a Large-scale Dataset for Fact Extraction and {VER}ification",
  author    = "Thorne, James and Vlachos, Andreas and Christodoulopoulos, Christos and Mittal, Arpit",
  booktitle = "Proceedings of the 2018 Conference of the North American Chapter of the Association for Computational Linguistics: Human Language Technologies, Volume 1 (Long Papers)",
  year      = "2018",
  month     = jun,
  address   = "New Orleans, Louisiana",
  publisher = "Association for Computational Linguistics",
  pages     = "809--819",
  doi       = "10.18653/v1/N18-1074",
  url       = "https://aclanthology.org/N18-1074/"
}

@inproceedings{gao-etal-2023-enabling,
  title     = "Enabling Large Language Models to Generate Text with Citations",
  author    = "Gao, Tianyu and Yen, Howard and Yu, Jiatong and Chen, Danqi",
  booktitle = "Proceedings of the 2023 Conference on Empirical Methods in Natural Language Processing",
  year      = "2023",
  month     = dec,
  address   = "Singapore",
  publisher = "Association for Computational Linguistics",
  pages     = "6465--6488",
  doi       = "10.18653/v1/2023.emnlp-main.398",
  url       = "https://aclanthology.org/2023.emnlp-main.398/"
}

\clearpage

\appendices

\section*{Supplementary Material}

The REASONS dataset is publicly available on Zenodo \url{(https://zenodo.org/records/18264033)}. The supplementary material provides prompt templates, domain-level evaluation results, model outputs, configuration files, retrieval and generation settings, and additional documentation required for reproduction.
\subsection*{A. Prompt Templates and Model Specifications}

\noindent
\colorbox{gray!10}{
\parbox{0.96\columnwidth}{
\textbf{Task T1: Author Attribution}

Given the following paper title, identify the complete list of
authors. If you are uncertain, output \texttt{PASS}.

\textbf{Input:} Paper title

\textbf{Output:} Author list or \texttt{PASS}
}}

\vspace{2mm}

\noindent
\colorbox{gray!10}{
\parbox{0.96\columnwidth}{
\textbf{Task T2: Title Attribution}

Given the following citation sentence, identify the cited
paper title. If you are uncertain, output \texttt{PASS}.

\textbf{Input:} Citation sentence

\textbf{Output:} Paper title or \texttt{PASS}
}}

\vspace{2mm}

\noindent
\colorbox{gray!10}{
\parbox{0.96\columnwidth}{
\textbf{Metadata-Augmented Prompt (T2)}

Given the following citation sentence and metadata,
identify the cited paper title.

If you are uncertain, output \texttt{PASS}.

\textbf{Sentence:}
\textit{{citation sentence}}

\textbf{Abstract:}
\textit{{paper abstract}}

\textbf{Authors:}
\textit{{author list}}

\textbf{Output:}
Paper title or \texttt{PASS}
}}

\noindent
\colorbox{gray!10}{
\parbox{0.96\columnwidth}{
\textbf{CMP Prompt (T2)}

Identify the cited paper title.

If uncertain, output \texttt{PASS}.

\textbf{Stage 1 Input:}
Sentence only

\textbf{Stage 2 Input:}
Sentence + Abstract

\textbf{Stage 3 Input:}
Sentence + Abstract + Authors

\textbf{Output:}
Paper title or \texttt{PASS}
}}

\subsection*{Model Specifications}



\begin{itemize}
\item \textbf{GPT-o1 (G1)}: OpenAI API with default decoding parameters.
\item \textbf{GPT-4o (G2)}: OpenAI API, temperature = 1.0, top-p = 1.0, max tokens = 256.
\item \textbf{GPT-3.5-Turbo (G3)}: OpenAI API, temperature = 1.0, top-p = 1.0, max tokens = 256.
\item \textbf{pplx-7b-chat (P)}: evaluated through the Perplexity API.
\item \textbf{DeepSeek-R1 (D)}: greedy decoding (\texttt{do\_sample=False}).
\item \textbf{Mistral-7B-v0.2-Instruct (M)}: temperature = 1.0, top-p = 0.95, max new tokens = 300.
\item \textbf{Llama-3.1-8B-Chat (L)}: temperature = 1.0, top-p = 0.95, max new tokens = 300.
\end{itemize}

\begin{table}[h]
\centering
\small
\begin{tabular}{ll}
\hline
Component & Model \\
\hline
Embedding Model & BAAI/bge-small-en-v1.5 \\
Cross Encoder & ms-marco-MiniLM-L-12-v2 \\
Llama Tokenizer & NousResearch/Llama-3.1-8b-chat-hf \\
Mistral Tokenizer & mistralai/Mistral-7B-v0.2 \\
\hline
\end{tabular}
\caption{Retrieval and tokenization components.}
\label{imp_det}
\end{table}

Experiments were conducted on NVIDIA A100 and L40S GPUs. Resources will be released upon publication.
For zero-context, metadata-augmented, and CMP experiments,
API-based models (GPT-4o, GPT-3.5, Perplexity) were evaluated
using their standard sampling configuration (temperature=1.0,
top-p=1.0). 
For RAG-based experiments using Mistral and Llama models,
generation employed stochastic decoding (temperature=1.0,
top-p=0.95, max\_new\_tokens=300). For DeepSeek adversarial and retrieval experiments, greedy
decoding (do\_sample=False) was used to ensure reproducible
attribution behavior under controlled retrieval conditions.

All implementation details, including generation parameters (temperature, max tokens, top-p), embedding models, retrieval hyperparameters, and top-$k$ values, are provided in Suppl. Table~\ref{imp_det}. The cost associated with this research includes expenses for utilizing OpenAI API, totaling \$640.37. Additionally, the use of \texttt{Perplexity} API incurred costs amounting to \$259.39. Furthermore, GPU resources, we used Replicate\footnote{\url{https://replicate.com/}} API for our experiments, amounted to \$466.22. For dataset creation, we used Oxylab for \$249 for a month. In total, the expenses for APIs were \$1614.98.

\textbf{Quality Control and Challenging Instances: }We validate citation instances through exact matching between in-text citation strings and verified bibliographic metadata, removing unmatched or malformed entries. However, we intentionally retain citation contexts that are fragmented, numeric-heavy, or difficult to interpret in isolation. Such cases naturally occur in scientific writing and represent realistic attribution challenges. Preserving these difficult examples enables a more rigorous evaluation of abstention and hallucination behavior, rather than restricting the benchmark to only clean, self-contained citation contexts.

\subsection*{B. Adversarial attribution setting:}
 To examine how attribution behavior changes when credibility cues are misleading, we introduce an adversarial evaluation setting. In this setting, the attribution task remains unchanged, but the metadata provided to the model is deliberately perturbed to be plausible yet incorrect. This allows us to test whether systems rely on surface-level similarity rather than contextual alignment when justifying sources. Let $x$ denote the query input (paper title for Task~T1 or cited sentence for Task~T2), and let $m$ denote the correct citation metadata. Under standard conditions, the model produces an attribution $ \hat{c} = f(x, m).$ For adversarial evaluation, we replace $m$ with perturbed metadata $\tilde{m}$ such that $\tilde{m} \neq m$ but remains highly similar: $ \hat{c}^{\,adv} = f(x, \tilde{m}),  \text{where } \mathrm{sim}(m, \tilde{m}) \ge \tau. $
Here, $\mathrm{sim}(\cdot,\cdot)$ denotes a similarity function and $\tau \geq 0.85$ is a threshold chosen to ensure substitutions are realistic and difficult to distinguish based on surface cues alone. For Task~T1, adversarial inputs are constructed by replacing the correct paper title with a similar but incorrect title. For Task~T2, adversarial inputs are created by substituting either the cited paper title or abstract with a highly similar alternative. No additional training or fine-tuning is performed in this setting. This isolates attribution behavior under misleading evidence and allows us to observe whether systems abstain or confidently assert incorrect attribution.

\begin{table}[t!]
\centering
\footnotesize
\renewcommand{\arraystretch}{0.95}
\resizebox{\columnwidth}{!}{
\begin{tabular}{p{2.1cm} p{1.35cm} p{1.35cm} p{1.35cm}}
\toprule
\textbf{Group} 
& \textbf{AR (\%)} 
& \textbf{HR (\%)} 
& \textbf{BLEU} \\ 
\midrule

\multicolumn{4}{c}{\textbf{Adversarial Setting: Changing Paper Title {(Task T2)}}} \\
\midrule
GPT-o1 (reasoning) & 96.23 & \textbf{17.99} & \textbf{0.6210} \\
GPT-4o         & 31.45 & 83.66 & 0.0524 \\
GPT-3.5        & 68.55 & 87.35 & 0.0389 \\
D              & 89.31 & \ul{61.31} & 0.0178 \\
RAG(M)         & 3.14  & 86.78 & 0.0796 \\
RAG(L)         & 5.03  & 87.56 & 0.0628 \\
AdvRAG(L)      & \textbf{0.00}  & 85.72 & 0.1322 \\
AdvRAG(M)      & \textbf{0.00}  & 75.41 & \ul{0.1569} \\

\midrule
\multicolumn{4}{c}{\textbf{Adversarial Setting: Changing Paper Abstract (Task T2)}} \\
\midrule
GPT-o1 (reasoning) & 95.60 & \textbf{38.49} & \textbf{}0.4595 \\
GPT-4o         & 32.70 & 86.22 & 0.0396 \\
GPT-3.5        & 76.10 & 91.64 & 0.0034 \\
DeepSeek       & 93.08 & 94.72 & 0.0000 \\
RAG(M)         & 7.55  & 89.44 & 0.0520 \\
RAG(L)         & 2.52  & 90.16 & 0.0445 \\
AdvRAG(L)      & \textbf{0.00}  & 39.67 & 0.4101 \\
AdvRAG(M)      & \textbf{0.00}  & \ul{39.57} & \ul{0.4904} \\

\bottomrule
\end{tabular}
}
\caption{Attribution behavior under adversarial metadata in Task~T2. Either the cited paper title or abstract is replaced with a highly similar but incorrect alternative. Adversarial metadata substantially increases hallucination while suppressing abstention, revealing a tendency to assert sources under misleading credibility cues.}
\label{tab:adv_metadata_t2}
\end{table}

 \noindent \textit{\textbf{Attribution behavior under adversarial metadata:}} 
 When titles/abstracts are replaced with highly similar but incorrect alternatives, HR increases sharply and AR often decreases (Table~\ref{tab:adv_metadata_t2}). Some proprietary models increase AR, but retrieval-driven systems commonly continue to answer. Models overweight semantic proximity and ``credibility signals'' (paper-like text), treating them as confirmation. RAG pipelines can inadvertently validate the wrong candidate by retrieving a coherent-looking neighbor document. This is not just a worst-case attack; it demonstrates a common real-world failure: near-duplicate literature and ambiguous citations can trigger confident misattribution.

\subsection*{C. Human Evaluation}
\noindent \textbf{Analysis of paraphrase cases.} Among the 500 non-pass instances evaluated, only 26 (5.2\%) involved semantically correct paraphrases that automated metrics incorrectly penalized (e.g., ``Reliable Attribution in Scientific Language Models'' vs. ``Trustworthy Attribution for Scholarly Language Models'', author name variations like ``A. Vaswani'' vs. ``Ashish Vaswani''). Conversely, 330 instances (66.0\%) were factually incorrect hallucinations that both automated metrics and human experts identified (wrong paper, wrong authors, fabricated titles). This 12.7:1 ratio (330:26) confirms that hallucination is predominantly a factual error problem, not a paraphrase-detection problem, thereby validating our automated measurement approach. The remaining 144 instances (28.8\%) were judged as ``Partially Correct'' by human annotators—cases where models identified the correct general paper but with incomplete or slightly inaccurate information (e.g., missing co-authors, abbreviated journal names, minor title variations that changed meaning). Our automated HR metric conservatively counts these as errors, accounting for the slight divergence between automated and human-judged error rates in Table~\ref{tab:human_validation}.

 \noindent \textbf{Abstention appropriateness findings.} Across the 500 pass instances, human experts judged 271 abstentions (54.2\%) as appropriate (the model correctly declined when evidence was insufficient, e.g., zero-context ambiguous queries) and 229 (45.8\%) as unnecessary (the model could have answered given available evidence, e.g., when complete metadata was provided). This near-even split indicates that current LLMs miscalibrate abstention decisions, oscillating between over-caution when evidence suffices and answering when abstention is warranted. It also motivates evaluating AR and HR jointly rather than optimizing either in isolation: very low AR can yield confident but incorrect answers (epistemic risk), while very high AR can lead to unnecessary abstentions (reduced utility).
\end{document}